\documentclass[10pt,twocolumn,letterpaper]{article}

\usepackage{wacv}              
\usepackage{multirow}
\usepackage{float}
\usepackage{stfloats}
\usepackage{graphicx}
\usepackage{xcolor}
\usepackage{colortbl}
\usepackage{booktabs}
\usepackage{svg}        
\usepackage{subcaption} 
\usepackage{caption}
\usepackage{balance}
\newcommand{\mName}{DnR\xspace}

\definecolor{wacvblue}{rgb}{0.21,0.49,0.74}

\usepackage[pagebackref,breaklinks,colorlinks,allcolors=wacvblue]{hyperref}

\def\wacvPaperID{3338} 
\def\confName{WACV}
\def\confYear{2026}

\title{Divide and Refine: Enhancing Multimodal Representation and Explainability \\ for Emotion Recognition in Conversation} 

\author{
Anh-Tuan Mai$^{1,2}$\\
\and
Cam-Van Thi Nguyen$^{1}$\\
\and
Duc-Trong Le$^{1}$\thanks{Corresponding author}\\
\and
$^1$VNU University of Engineering and Technology \\
$^2$FPT Software AI Center
\\
{\tt\small \{24025161, vanntc, trongld\}@vnu.edu.vn}
}

\begin{document}
\maketitle
\begin{abstract}
Multimodal emotion recognition in conversation (MERC) requires representations that effectively integrate signals from multiple modalities. These signals include modality-specific cues, information shared across modalities, and interactions that emerge only when modalities are combined. In information-theoretic terms, these correspond to \emph{unique}, \emph{redundant}, and \emph{synergistic} contributions. An ideal representation should leverage all three, yet achieving such balance remains challenging. Recent advances in contrastive learning and augmentation-based methods have made progress, but they often overlook the role of data preparation in preserving these components. In particular, applying augmentations directly to raw inputs or fused embeddings can blur the boundaries between modality-unique and cross-modal signals.  
To address this challenge, we propose a two-phase framework \emph{\textbf{D}ivide and \textbf{R}efine} (\textbf{DnR}). In the \textbf{Divide} phase, each modality is explicitly decomposed into uniqueness, pairwise redundancy, and synergy. In the \textbf{Refine} phase, tailored objectives enhance the informativeness of these components while maintaining their distinct roles. The refined representations are plug-and-play compatible with diverse multimodal pipelines.  
Extensive experiments on IEMOCAP and MELD demonstrate consistent improvements across multiple MERC backbones. These results highlight the effectiveness of explicitly dividing, refining, and recombining multimodal representations as a principled strategy for advancing emotion recognition.
Our implementation is available at \url{https://github.com/mattam301/DnR-WACV2026}

\end{abstract}
    

\begin{figure}
    \centering
    \includegraphics[width=\linewidth]{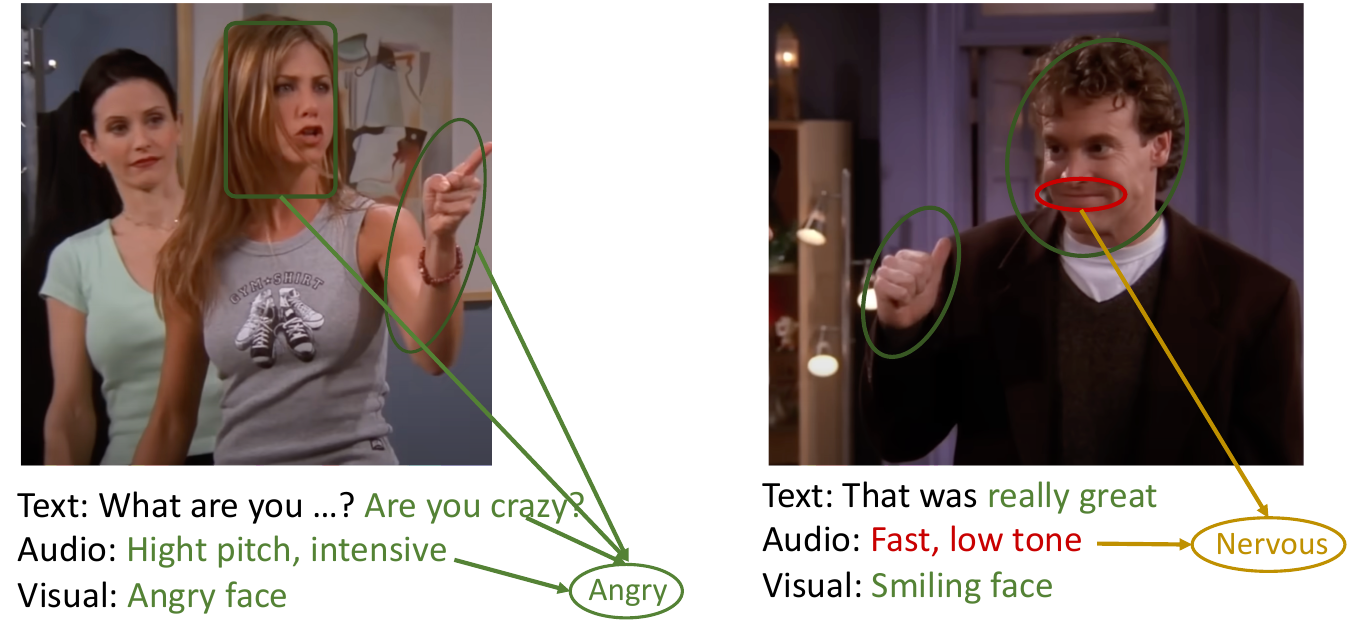}
    \caption{Illustrative examples of modality interactions in MER. \textbf{Left}: verbal, acoustic, and visual cues consistently convey anger, reflecting redundancy across modalities. \textbf{Right}: text and visual cues redundantly suggest positivity, yet the unique acoustic cue (\textit{fast, low tone}) provides critical evidence which, when integrated synergistically, reveals the true emotion of nervousness. These cases demonstrate that unique information, though often overshadowed by redundancy, plays a vital role in synergy and accurate emotion recognition.}
    \label{fig:motivation}
\end{figure}

\section{Introduction}
\label{sec:intro}
Multimodal machine learning seeks to integrate information from heterogeneous data sources \cite{baltruvsaitis2018multimodal, liang2024foundations, nguyen-etal-2024-curriculum}, with the goal of constructing representations that preserve complementary and overlapping signals for downstream modeling. In multimodal emotion recognition in conversation (MERC), this involves leveraging audio, visual, and textual cues, each offering distinct perspectives on affective states \cite{ahmed2023systematic, pan2023review}. Prior work has explored multimodal representations through graph- or attention-based fusion \cite{hu-etal-2021-mmgcn, joshi2022cogmen, nguyen2023conversation} and contrastive alignment in shared spaces \cite{song2025leveraging}, achieving strong results but oversimplifying modality interactions. In reality, these interactions are nuanced: some cues are modality-specific, others are redundantly available, and still others emerge only through joint integration \cite{williams2010nonnegative, zadeh2018memory, mai2019divide}. In information-theoretic terms, these correspond to \emph{unique}, \emph{redundant}, and \emph{synergistic} contributions~\cite{liang2023quantifying}. Fusion-based models risk overemphasizing redundancy, while graph-based models oversimplify dependencies. Recent advances in Partial Information Decomposition (PID) \cite{williams2010nonnegative, bertschinger2014quantifying, liang2023quantifying, dufumier2024align} provide a principled framework to distinguish these contributions, offering a finer-grained view of multimodal interactions beyond simple feature fusion or connectivity.

In MERC, these distinctions are intuitive. Unique information ($U$) arises when only one modality provides a discriminative cue, such as acoustic prosody indicating sarcasm despite neutral text. Redundant information ($R$) corresponds to cues independently available from multiple modalities, for example, both a smiling face and a cheerful voice signaling happiness. Synergistic information ($S$) emerges only through joint integration, as in irony, which requires reconciling positive lexical content with a negative vocal tone. Figure~\ref{fig:motivation} illustrates these cases: the left example reflects redundancy across modalities, whereas the right shows how a unique acoustic cue, when combined synergistically, reveals the correct emotion. These examples underscore that $U$, $R$, and $S$ are not theoretical abstractions but interpretable roles in multimodal emotion understanding.  

Modeling these factors remains challenging. Fusion strategies often entangle $U$, $R$, and $S$, leading to degenerate solutions where redundant signals dominate while unique and synergistic cues are suppressed. Contrastive methods such as CoMM \cite{dufumier2024align} attempt to emphasize underutilized signals through augmentation, but they rely on crafted manipulations of inputs that may distort high-level emotional cues. For example, changing the color of a flower image is a valid augmentation for object recognition, whereas applying a similar transformation to an image of a laughing person may corrupt the abstract and semantically fragile cue of a smile. Factorization-based methods such as SMURF \cite{wortwein2024smurf} highlight the benefits of separating unique and redundant subspaces, but they neither enhance the informativeness of shared representations nor explicitly capture synergy.

To address these limitations, we propose a two-phase framework based on the principle of \textbf{Divide and Refine - \mName{}}. In the first phase (\textbf{Divide}), modality representations are disentangled into uniqueness, redundancy, and synergy, yielding structured features that explicitly separate informational roles. In the second phase (\textbf{Refine}), targeted objectives are applied: redundancy is augmented in representation space to improve robustness, while uniqueness and synergy are preserved to maintain distinct contributions. This design enhances the informativeness of multimodal features and remains compatible with diverse MER backbones.  

Our main contributions are threefold:
\begin{itemize}
    \item We introduce a unified formulation of multimodal representations for MERC grounded in Partial Information Decomposition, where the \textbf{Divide} phase disentangles modality features into uniqueness, redundancy, and synergy. 
    \item We propose a redundancy-focused augmentation and refinement strategy, forming the \textbf{Refine} phase, which strengthens shared information while preserving the distinct contributions of unique and synergistic signals. 
    \item We integrate the proposed \textbf{Divide and Refine - \mName{}} framework as a model-agnostic module and demonstrate consistent improvements across multiple MER backbones on IEMOCAP and MELD, validating its generality and effectiveness.
\end{itemize}

\section{Related Work}

\subsection{Multimodal Emotion Recognition}  
Multimodal emotion recognition in conversation (MERC) aims to classify utterance-level emotions by jointly leveraging speech, text, and visual modalities.
Early studies primarily focused on textual context, with models such as DialogueGCN~\cite{ghosal2019dialoguegcn}, COSMIC~\cite{ghosal2020cosmic}, and DialogueCRN~\cite{hu2021dialoguecrn} employing sequential or graph-based architectures to capture interaction flow. More recently, transformer-based approaches like DialogXL~\cite{shen2021dialogxl} incorporated pretrained language models to enhance context modeling.  
To integrate multimodal signals, fusion-based methods combine features across modalities. Approaches such as bc-LSTM~\cite{tao2021short} and CMN~\cite{hazarika2018conversational} adopt temporal modeling or simple feature concatenation, while more advanced designs such as SMIN~\cite{lian2022smin} and SDT~\cite{ma2023transformer} refine fusion through semi-supervised learning and transformer-based hierarchical integration. However, these approaches often allow dominant modalities to overshadow subtle yet informative cues, making them sensitive to modality imbalance. Graph-based methods attempt to address this limitation by explicitly encoding conversational and multimodal structures. MMGCN~\cite{hu-etal-2021-mmgcn} showed the benefits of undirected graphs for multimodal integration, while COGMEN~\cite{joshi2022cogmen} and CORECT~\cite{nguyen2023conversation} introduced structured conversation graphs with enhanced interaction modeling. More recently, GraphSmile~\cite{li2025tracing} unified emotion recognition and sentiment analysis through multi-task optimization.  

Despite these advances, most existing approaches emphasize architectural design while overlooking the informational roles carried by each modality. Multimodal signals may contribute uniquely, redundantly, or synergistically, and failing to disentangle these factors limits the expressiveness of learned representations. This motivates our representation-centric perspective, where we employ Partial Information Decomposition (PID) as a principled tool to divide and refine the distinct contributions of each modality.



\subsection{Partial Information Decomposition in Multimodal Learning}

Recent work has underscored that multimodal representations are not uniform entities but can be decomposed into distinct informational roles, namely uniqueness, redundancy, and synergy \cite{williams2010nonnegative}. This perspective has inspired approaches that attempt to isolate or rebalance these components during learning~\cite{bertschinger2014quantifying}. For instance, CoMM~\cite{dufumier2024align} employs contrastive objectives and augmentation strategies to mitigate redundancy dominance in fused representations, thereby encouraging attention to modality-specific cues. SMURF~\cite{wortwein2024smurf} further shows that explicitly disentangling unique and redundant contributions yields more interpretable and robust predictions. More recently, FactorCL~\cite{liang2023factorized} incorporates uniqueness and redundancy into multimodal contrastive learning, yet its reliance on conditional augmentations leaves synergistic information underexplored. Collectively, these studies highlight the value of decomposition for both interpretability and performance, while also revealing the need for unified frameworks that explicitly account for all three PID components.



\section{Preliminaries}
We briefly review the theoretical foundations of our approach. We first introduce Partial Information Decomposition (PID), which formalizes how modalities contribute to a target variable, and then outline the multimodal learning setting with emphasis on MERC.

\subsection{Partial Information Decomposition}

A key challenge in multimodal learning is to characterize how modalities contribute to predicting a target variable. The total mutual information $I(Y; M_1, M_2, \dots, M_K)$ quantifies the overall reduction in uncertainty about $Y$, but conflates distinct contributions. It is therefore essential to distinguish whether information is provided uniquely by one modality, redundantly by several, or synergistically through their interaction.

\paragraph{PID framework.}  
Partial Information Decomposition (PID)~\cite{williams2010nonnegative, ince2017measuring} provides a principled framework for disentangling multimodal contributions. For the two-modality case $(M_1, M_2)$, PID decomposes the joint mutual information into four disjoint components:
\begin{equation}
I(Y; M_1, M_2) = U_1 + U_2 + R + S,
\end{equation}
where $U_1$ and $U_2$ denote the unique information about $Y$ accessible only from $M_1$ or $M_2$, respectively; $R$ denotes the redundant information that is independently available from both modalities; and $S$ represents the synergistic information that emerges only when $M_1$ and $M_2$ are jointly considered. This decomposition generalizes to $K$ modalities through a lattice structure of informational atoms, with each atom corresponding to unique, redundant, or synergistic contributions associated with specific modality subsets~\cite{bertschinger2014quantifying}. PID thus enables a fine-grained analysis of cross-modal interactions beyond what is revealed by mutual information alone.


\subsection{Multimodal Emotion Recognition}


In MERC, each utterance is represented across $M$ modalities, typically audio $(a)$, text $(t)$, and vision $(v)$.  
Formally, for the $j$-th utterance in conversation $U_i$, we denote
\begin{equation}
    x_{ij} = \{x^a_{ij}, \; x^t_{ij}, \; x^v_{ij}\},
\end{equation}
where $x^m_{ij} \in \mathbb{R}^{d_m}$ is the feature vector of modality $m$. The task is to learn a prediction function
\begin{equation}
    F: \{x^a_{ij}, x^t_{ij}, x^v_{ij}\} \mapsto \hat{y}_{ij} \in \mathcal{C},
\end{equation}
that assigns each utterance an emotion label $\hat{y}_{ij}$ from a predefined category set $\mathcal{C}$, conditioned on multimodal conversational context. Existing fusion- and graph-based models design $F(\cdot)$ to integrate multimodal features but often conflate different types of cross-modal information. From the perspective of Partial Information Decomposition (PID), each modality contributes \emph{unique}, \emph{redundant}, and \emph{synergistic} signals, yet standard approaches rarely separate these roles. This can result in redundancy dominating or synergy being overlooked.  

\begin{figure*}
    \centering
    \includegraphics[width=0.7\linewidth]{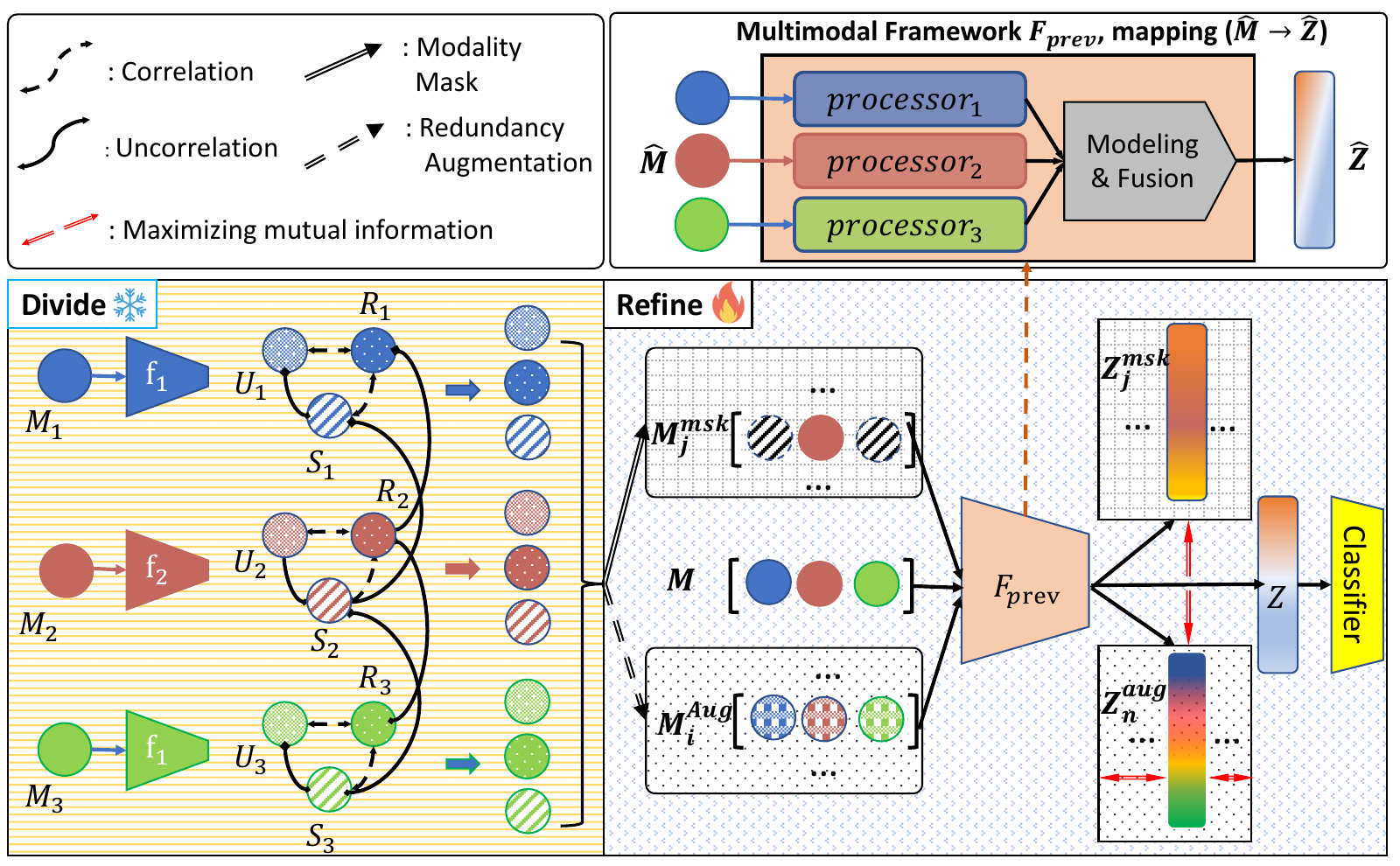}
    \caption{Overall Framework of \textbf{DnR}, with two consecutive phases: \textbf{Divide} and \textbf{Refine}. \textbf{Divide} phase is conducted first as all outputs of this phase are concatenated to make predict to classification task with ${L_{corr}}$ and ${L_{uncor}}$ as auxiliary for separating uniqueness, redundancy and synergy. All trainable parameters of this phase is frozen before \textbf{Refine} phase}
    \label{fig:overall}
\end{figure*}
\section{Methodology}
\label{sec:method}
Figure \ref{fig:overall} present our end-to-end plug-and-play framework, designed to enhance multimodal representations and predictions while being seamlessly compatible with diverse MERC backbones.

\subsection{Divide: Label-Aware Multi-Modality Decomposition}

\paragraph{Representation decomposition.}  
Given a multimodal backbone encoder with per-modality feature extractors $f_m(\cdot)$, we extend its outputs by explicitly decomposing each modality into three informational streams. For an input $x_m$ from modality $m$, the encoder produces the modality-specific representation $h_m$ as:
\begin{equation}
    h_m = f_m(x_m) = \Big(h_m^U, \; h_m^R, \; h_m^S\Big),
\end{equation}
where $h_m \in \mathbb{R}^{d_m}$, $d_m$ is the output dimention of the encoder $f_m$. $h_m^U \in \mathbb{R}^{d_m}$ denotes the \emph{unique} embedding capturing modality-specific cues, $h_m^R \in \mathbb{R}^{d_m}$ is a \emph{redundant} embedding aligned with overlapping information across modalities, and $h_m^S \in \mathbb{R}^{d_m}$ contributes to the global \emph{synergy} embedding that captures higher-order cross-modal interactions. Thus, instead of a single fused representation, each modality is structured into complementary components that can be refined in the subsequent phase.  

\paragraph{Training objectives.}  
The Divide module is optimized jointly with the backbone through the task loss and two regularization terms:  
\begin{equation}
\mathcal{L}_{\text{Divide}} = \mathcal{L}_{\text{task}} 
+ \lambda_{\text{uncor}} \, \mathcal{L}_{\text{uncor}}
+ \lambda_{\text{corr}} \, \mathcal{L}_{\text{corr}}.
\end{equation}
where $\lambda_{\text{uncor}}, \lambda_{\text{corr}} \in [0, 1]$ are hyper-parameters  

\textbf{Task loss.}  
All components are supervised by the MERC backbone. A lightweight predictor $g(\cdot)$ maps each embedding to logits, and the final output is obtained by additive aggregation:
\begin{equation}
    \hat{y} = \sum_m g(h_m^U) \;+\; \sum_m g(h_{m}^R) \;+\; \sum_m g(h_m^S).
\end{equation}

\textbf{Uncorrelation loss.}  
To avoid redundancy collapsing into uniqueness, we penalize their correlation within each modality:
\begin{equation}
    \mathcal{L}_{\text{uncor}} = \sum_m \big|\text{corr}(h_m^U, h_{m}^R)\big|.
\end{equation}

\textbf{Correlation-enhancing loss.}  
We encourage consistency across redundant streams, enforce alignment across synergy streams, and promote synergy–uniqueness coupling, which ensures that synergy captures complementary interactions rather than duplicating redundancy:
\begin{align}
    \mathcal{L}_{\text{corr}} = 
    &- \sum_{m \neq s} \text{corr}\!\big(h_{m}^R, h_{s}^R\big) 
    - \sum_{m \neq s} \text{corr}\!\big(h_{m}^S, h_{s}^S\big) \nonumber \\
    &- \alpha \sum_m \text{corr}(h_m^S, h_m^U).
\end{align}

By design, this decomposition module can be plugged into arbitrary MERC backbones, restructuring their representations into $U$, $R$, and $S$ streams while leaving the backbone architecture and objective intact. 

\subsection{Refine: Redundancy-Augmented Refinement}

The refine phase focuses on redundancy augmentation. Unique information ($U$) must be preserved, and synergistic information ($S$) is fragile to perturbation, whereas redundancy ($R$) offers a robust locus for augmentation. From the perspective of Partial Information Decomposition (PID), this strategy perturbs $R$ while preserving $U$ and $S$, thereby enforcing more informative and robust multimodal representations. The formulation is given as follows:

\begin{equation}
    I(Y; M_1, M_2, \dots, M_K) = \sum_{m} U_m + R + S,
\end{equation}
where $U_m$, $R$, and $S$ denote unique, redundant, and synergistic contributions, respectively. A contrastive objective maximizes the mutual information between original and augmented representations, $I(Z; Z')$, where $Z$ and $Z'$ are two correlated views of the multimodal input. If augmentations are applied indiscriminately at the raw input level, they may reduce $U$ or distort $S$. Refinement via redundancy augmentation averts this while preserving useful information.

\paragraph{Formulation.} 
Let $F_{\text{prev}}$ denote an arbitrary multimodal backbone (e.g., MMGCN, MM-DFN) into which our refined representations are fed.
For a training sample $i$, the decomposition phase outputs a set of modality-specific components, which can be written as:
\begin{equation}
    \mathcal{M}_{(i)} = \big\{\, h_m^U(i)\oplus\; h_m^R(i) \oplus\; h_m^S(i) \;\big|\; \forall m \,\big\},
\end{equation}
where $h_m^U(i)$, $h_m^R(i)$, and $h_m^S(i)$ denote the unique, redundant, and synergy embeddings of modality $m$, respectively. The symbol $\oplus$ stands for concatenation operation. From now on, they are denoted as $h_m^U,h_m^R,h_m^S$ for short.

Based on $\mathcal{M}_{(i)}$, we construct three sets of input:
\begin{equation}
\begin{aligned}
\mathcal{M} &= \big\{\, h_m^U(i)\oplus\; h_m^R(i)\oplus\; h_m^S(i) \;\big|\; \forall m \,\big\}, \\[4pt]
\mathcal{M}_{m} &= \big\{\, h_m^U(i)\oplus\; h_m^R(i)\oplus\; h_m^S(i)\,\big\} 
   \cup \big\{\, \mathbf{0} \;|\; \forall m' \neq m \,\big\}, \\[4pt]
\mathcal{M}_{aug}^{k} &= \Big\{\,
\{\, h_m^U(i)\oplus\; \tilde{h}_m^R(i,k)\oplus\; h_m^S(i) \;\big|\; \forall m \,\}
\;\},
\end{aligned}
\end{equation}
where $\mathcal{M}$ is the original multimodal bundle; $\mathcal{M}_{m}$ denotes the set of $m$ modality-specific version in which all modalities except $m$ are set to $\mathbf{0}$; and $\mathcal{M}_{aug}^{k}$ is one of $K$ augmented versions ($ k = 1,\dots,K$), in which each representation has redundancy vector $h_m^R(i)$ is replaced by its perturbed form $\tilde{h}_m^R(i) = \mathrm{Aug}(h_m^R(i))$.

All inputs are then passed into an arbitrary MERC backbone $F_{\text{prev}}$, yielding final fused representations:
\begin{equation}
    Z = F_{\text{prev}}(\mathcal{M}),
\end{equation}
where $\mathcal{M}$ denotes a set of multimodal representations; $F_{\text{prev}}$ accepts vanilla, masked, and augmented versions as input.






\paragraph{Contrastive objectives.}  
We adopt complementary InfoNCE losses to enhance redundancy robustness. For a minibatch of size $N$, let $\ell_{\text{InfoNCE}}(a,b)$ denote the symmetric bidirectional InfoNCE loss between two sets of embeddings $a,b \in \mathbb{R}^{N \times d}$.  

First, we enforce \emph{augmentation consistency} by requiring two stochastically augmented bundles to produce similar fused embeddings:
\begin{equation}
    \mathcal{L}_{\text{aug-intra}} = \sum_{k \neq n}\ell_{\text{InfoNCE}}(Z_{aug}^k, Z_{aug}^n).
\end{equation}
This objective encourages the backbone to focus on overall context rather than creating shortcut through over-dominated redundanct cues.

Second, we introduce an \emph{augmented–masked alignment} objective, aligning redundancy-augmented fused embeddings with masked fused embeddings:
\begin{equation}
    \mathcal{L}_{\text{aug-mask}} = \sum_{m \in \{a,v,t\}}\sum_{k \in K} \ell_{\text{InfoNCE}}(Z_{aug}^k, Z_m).
\end{equation}
This maximize the mutual information between augmented versions with each modal-specific representation, enhance the ability to extract uniqueness of each modality.

\paragraph{Overall Refine objective.}  
The overall refine objective combines the backbone’s supervised loss with the above contrastive terms:
\begin{equation}
    \mathcal{L}_{\text{Refine}} = \mathcal{L}_{\text{task}} + \lambda_{1}\mathcal{L}_{\text{aug-intra}} + \lambda_{2}\mathcal{L}_{\text{aug-mask}}.
\end{equation}

\paragraph{Compatibility.}  
Since $F_{\text{prev}}$ is an arbitrary MERC backbone, refinement requires no architectural modification. The only changes are (i) decomposing modality embeddings into $U$, $R$, and $S$ streams, (ii) applying redundancy augmentation, and (iii) adding auxiliary contrastive losses during training. This makes the refine phase a lightweight plug-in applicable to diverse MERC pipelines.

\begin{table}[t]
    \centering
    \caption{Dataset statistics}
    \label{tab:data-stats}
    \resizebox{0.85\columnwidth}{!}{%
\begin{tabular}{c|ccc|ccl}
\hline
\multirow{2}{*}{\textbf{Dataset}}                                  & \multicolumn{3}{c|}{\textbf{Dialogues}}                                 & \multicolumn{3}{c}{\textbf{Utterances}}                                  \\ \cline{2-7} 
                                                          & \multicolumn{1}{c|}{train} & \multicolumn{1}{c|}{valid} & test & \multicolumn{1}{c|}{train}  & \multicolumn{1}{c|}{valid} & test  \\ \hline
\begin{tabular}[c]{@{}c@{}}IEMOCAP\end{tabular} & \multicolumn{1}{c|}{108}   & \multicolumn{1}{c|}{12}    & 31   & \multicolumn{1}{c|}{5,146}  & \multicolumn{1}{c|}{664}   & 1,623 \\
MELD                                                     & \multicolumn{1}{c|}{1039} & \multicolumn{1}{c|}{114}   & 280  & \multicolumn{1}{c|}{9989} & \multicolumn{1}{c|}{1109} &  2610\\ \hline
\end{tabular}%
}
\end{table}
\begin{table}[t!]
\caption{Summary of multimodal feature extraction.}
\label{tab:feature_extraction}
\centering
\renewcommand{\arraystretch}{1}
\resizebox{0.85\columnwidth}{!}{%
\begin{tabular}{c|c|c|c}
\hline
\textbf{Dataset} & \textbf{Modality} & \textbf{Source} & \textbf{Dim} \\
\hline
\multirow{3}{*}{\textbf{IEMOCAP}} 
    & Acoustic & OpenSmile~\cite{eyben2010opensmile} & 100 \\
    & Visual & OpenFace~\cite{baltrusaitis2018openface} & 512 \\
    & Textual & sBERT~\cite{reimers-gurevych-2019-sentence} & 768 \\
\hline
\multirow{3}{*}{\textbf{MELD}} 
    & Acoustic & wav2vec~\cite{schneider2019wav2vec} & 512 \\
    & Visual & MA-Net~\cite{zhao2021learning} + MTCNN~\cite{zhang2016joint} & 1024 \\
    & Textual & DeBERTa~\cite{he2006deberta} & 1024 \\
\hline
\end{tabular}%
}
\end{table}
\section{Experiments}

\begin{table*}[!htbp]
\centering
\caption{Performance comparison on IEMOCAP dataset.}
\label{tab:iemocap-result}
\resizebox{0.75\textwidth}{!}{%
\begin{tabular}{lcccccccc}
\toprule
\multirow{2}{*}{Backbone} & \multicolumn{2}{c}{atv} & \multicolumn{2}{c}{av} & \multicolumn{2}{c}{at} & \multicolumn{2}{c}{tv} \\
\cmidrule(lr){2-3} \cmidrule(lr){4-5} \cmidrule(lr){6-7} \cmidrule(lr){8-9}
 & Acc & W-F1 & Acc & W-F1 & Acc & W-F1 & Acc & W-F1 \\
\midrule
MMGCN \cite{hu-etal-2021-mmgcn}       & 65.82 & 66.70 & 51.69 & 50.82 & 63.89 & 63.64 & 62.42 & 62.24 \\
+ \textbf{\mName{}}     & \textbf{66.90} & \textbf{67.96} (\( +1.26\)) & \textbf{53.05} & \textbf{52.08} (\( +1.26\)) & \textbf{65.06} & \textbf{65.42} (\( +1.78\)) & \textbf{63.28} & \textbf{63.37} (\( +1.13\)) \\
\midrule
DialogueGCN \cite{ghosal2019dialoguegcn} & 65.87 & 66.01 & 52.66 & 52.14 & 66.36 & 65.97 & 65.25 & 64.80 \\
+ \textbf{\mName{}}     & \textbf{67.78} & \textbf{67.91} (\( +1.90\)) & \textbf{56.25} & \textbf{56.25} (\( +4.11\)) & \textbf{68.27} & \textbf{68.11} & 65.44 & 65.92 (\( +1.12\))\\
\midrule
MM-DFN \cite{hu2022mm}      & 65.19 & 65.20 & 50.40 & 48.78 & 63.89 & 62.46 & 60.20 & 60.14 \\
+ \textbf{\mName{}}     & \textbf{66.30} & \textbf{66.51} (\( +1.31\)) & \textbf{57.73} & \textbf{57.48} (\( +8.70\)) & \textbf{64.14} & \textbf{64.28} (\( +1.82\)) & \textbf{63.34} & \textbf{63.83} (\( +3.69\)) \\
\midrule
GraphSmile \cite{li2025tracing}  & 69.29 & 69.19 & - & - & - & - & - & - \\
+ \textbf{\mName{}}     & \textbf{70.24} & \textbf{70.15} (\( +0.96\)) & - & - & - & - & - & - \\
\midrule
SDT \cite{ma2023transformer}         & 71.47 & 71.80 & 58.72 & 58.05 & 69.81 & 69.82 & 66.40 & 66.10 \\
+ \textbf{\mName{}}     & \textbf{72.83} & \textbf{73.13} (\( +1.33\)) & \textbf{60.20} & \textbf{60.03} (\( +1.98\)) & \textbf{70.49} & \textbf{70.81} (\( +1.70\)) & \textbf{68.20} & \textbf{67.90} (\( +1.80\)) \\
\bottomrule
\end{tabular}%
}
\end{table*}

\begin{table*}[!htbp]
\centering
\caption{Performance comparison on MELD dataset.}
\label{tab:meld-result}
\resizebox{0.75\textwidth}{!}{%
\begin{tabular}{lcccccccc}
\toprule
\multirow{2}{*}{Backbone} & \multicolumn{2}{c}{atv} & \multicolumn{2}{c}{av} & \multicolumn{2}{c}{at} & \multicolumn{2}{c}{tv} \\
\cmidrule(lr){2-3} \cmidrule(lr){4-5} \cmidrule(lr){6-7} \cmidrule(lr){8-9}
 & Acc & W-F1 & Acc & W-F1 & Acc & W-F1 & Acc & W-F1 \\
\midrule
MMGCN \cite{hu-etal-2021-mmgcn}       & 61.37 & 58.78 & 48.30 & 43.24 & 61.80 & 58.92 & 60.46 & 58.32 \\
+ \textbf{\mName{}}     & \textbf{61.46} & \textbf{59.94} (\( +1.16\)) & \textbf{48.20} & \textbf{43.63} (\( +0.39\)) & \textbf{61.07} & \textbf{59.79} (\( +0.87\)) & \textbf{61.69} & \textbf{59.46} (\( +1.14\)) \\
\midrule
DialogueGCN \cite{ghosal2019dialoguegcn} & 61.23 & 58.90 & 47.28 & 42.70 & 60.88 & 58.21 & 61.38 & 56.93 \\
+ \textbf{\mName{}}     & \textbf{61.80} & \textbf{59.64} (\( +0.74\)) & \textbf{48.97} & \textbf{43.12} (\( +0.42\)) & \textbf{60.92} & \textbf{58.65} (\( +0.44\)) & \textbf{61.38} & \textbf{58.81} (\( +1.88\)) \\
\midrule
MM-DFN \cite{hu2022mm}      & 60.50 & 58.11 & 46.13 & 41.45 & 61.83 & 58.54 & 62.07 & 58.44 \\
+ \textbf{\mName{}}     & 60.38 & \textbf{59.76} (\( +1.65\)) & \textbf{47.36} & \textbf{42.57} (\( +1.12\)) & {60.61} & \textbf{59.33} (\( +0.79\)) & \textbf{62.34} & \textbf{58.83} (\( +0.39\)) \\
\midrule
GraphSmile \cite{li2025tracing}  & 64.98 & 64.10 & - & - & - & - & - & - \\
+ \textbf{\mName{}}     & 65.60 & 64.85 (\( +0.75\)) & - & - & - & - & - & - \\
\midrule
SDT \cite{ma2023transformer}         & 68.12 & 64.28 & 56.74 & 49.10 & 67.85 & 62.83 & 67.92 & 61.92 \\
+ \textbf{\mName{}}     & \textbf{68.45} & \textbf{64.84} (\( +0.56\)) & \textbf{57.21} & \textbf{50.05} (\( +0.95\)) & \textbf{68.10} & \textbf{63.71} (\( +0.88\)) & \textbf{68.31} & \textbf{62.55} (\( +0.63\)) \\
\bottomrule
\end{tabular}%
}
\end{table*}

\subsection{Experimental Settings}

\textbf{Datasets.}  
We conduct experiments on three widely used multimodal emotion recognition benchmarks:  
\textbf{IEMOCAP} \cite{busso2008iemocap} is a dyadic conversation dataset annotated with categorical emotions such as happy, sad, neutral, and angry.  
\textbf{MELD} \cite{poria2018meld} consists of multiparty dialogues extracted from the TV series ``Friends'', annotated with speaker-level emotions.  
For each dataset, we adopt the standard preprocessing pipeline and train–test splits used in prior work. Table~\ref{tab:feature_extraction} summarizes the feature extraction process.
\textbf{Baselines.}  
To demonstrate the plug-and-play nature of our method, we integrate it with seven typical MER models:  
\textbf{MMGCN} \cite{hu-etal-2021-mmgcn},  
\textbf{DialogueGCN} \cite{ghosal2019dialoguegcn},  
\textbf{MM-DFN} \cite{hu2022mm},  
\textbf{SDT} (\cite{ma2023transformer}),  
\textbf{CORECT} \cite{nguyen2023conversation}
These backbones cover graph-based, attention-based, and dynamic fusion paradigms.

\textbf{Implementation details.}  
In Phase I, we pretrain modality encoders with decomposition heads for 30–80 epochs using AdamW with a learning rate of $1e^{-4}$. Phase II is trained for 20–50 epochs, applying representation-level augmentations only to the shared components. We set $\lambda_{\text{uncor}}=1.0$, $\lambda_{\text{corr}}=0.5$, and tune $\mu\in[0.5,2.0]$ by dataset. All models are implemented in PyTorch and trained on a single NVIDIA L4 GPU.

\textbf{Evaluation metrics.}  
We report weighted F1 and accuracy. For IEMOCAP, we follow the common six-class setting (hap, sad, neu, ang, exc, fru). 

\subsection{Main Results}

\textbf{IEMOCAP.}
Table~\ref{tab:iemocap-result} presents results on IEMOCAP under four modality settings, reporting accuracy and weighted F1. 
Across all backbone models, integrating \textbf{\mName{}} consistently yields performance gains. 
For example, MMGCN improves from 66.70 to 67.96 W-F1 in the full-modality \textit{atv} setting, while DialogueGCN shows an even larger gain (66.01 $\to$ 67.91). 
MM-DFN benefits most prominently in the reduced-modality \textit{av} case (48.78 $\to$ 57.48), highlighting the effectiveness of our method when fewer modalities are available. 
The strongest baseline, SDT, also achieves further gains, advancing from 71.80 to 73.13 W-F1 in \textit{atv}, with steady improvements across \textit{av}, \textit{at}, and \textit{tv}. 
Even GraphSmile, which already shows strong performance, observes consistent improvements. These results confirm that explicitly decomposing and refining multimodal signals provides benefits regardless of the underlying multimodal architecture. 
Notably, the consistent gains in partial-modality settings demonstrate that \textbf{\mName{}} improves robustness under missing or noisy modalities, improving both overall accuracy and F1.

\textbf{MELD}
Table~\ref{tab:meld-result} shows results on MELD. 
Across all backbone models, integrating \textbf{\mName{}} yields consistent improvements in both accuracy and weighted F1. 
For example, MMGCN improves from 58.78 to 59.94 W-F1 in the full-modality \textit{atv} setting, while DialogueGCN gains from 58.90 to 59.64.  
Even stronger baselines also benefit: GraphSmile increases by +0.75 W-F1, and SDT advances steadily from 64.28 to 64.84 in \textit{atv}, with further gains of +0.95 and +0.88 in \textit{av} and \textit{at}, respectively. 

Although the absolute gains on MELD are smaller compared to IEMOCAP, the uniform improvements across all backbones and modality subsets confirm the generality of our approach. 
Importantly, the gains in partial-modality settings further highlight \textbf{\mName{}}’s ability to refine redundancy and strengthen the use of limited cross-modal cues in emotion recognition task.

\begin{figure*}[t]
  \centering
  \resizebox{0.96\textwidth}{!}{ 
    \begin{tabular}{ccc} 
      \includegraphics[width=\linewidth]{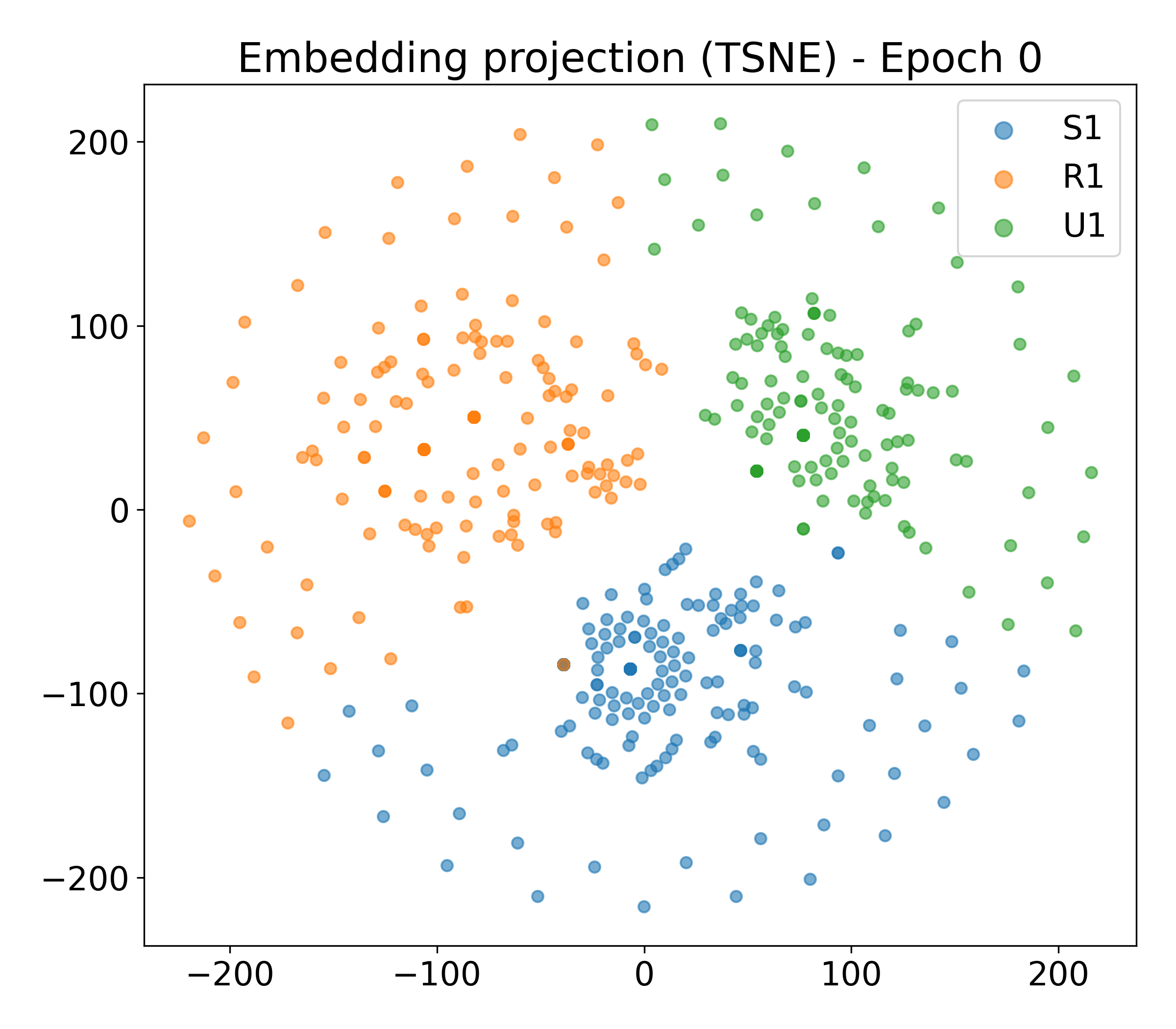} &
      \includegraphics[width=\linewidth]{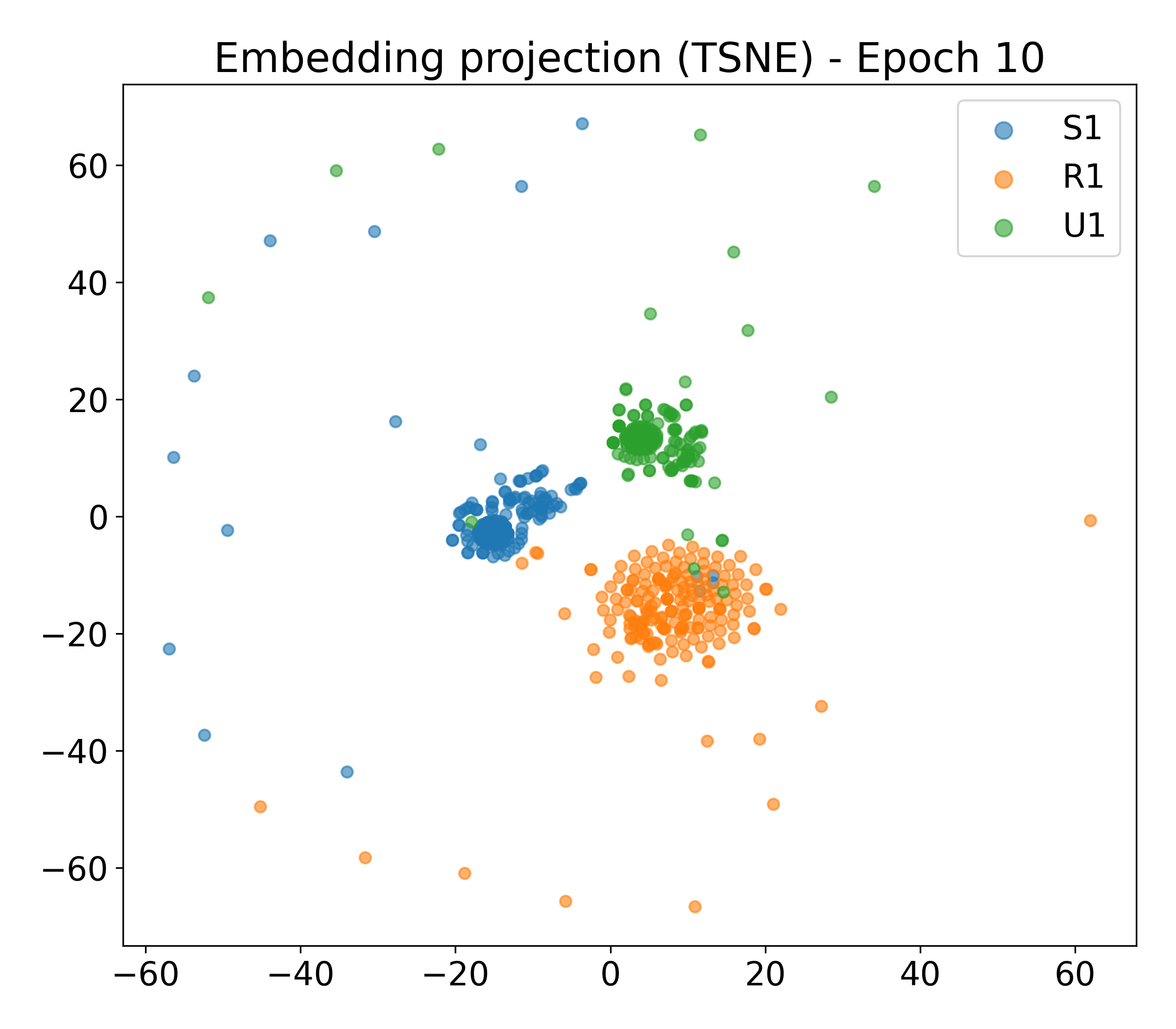} &
      \includegraphics[width=\linewidth]{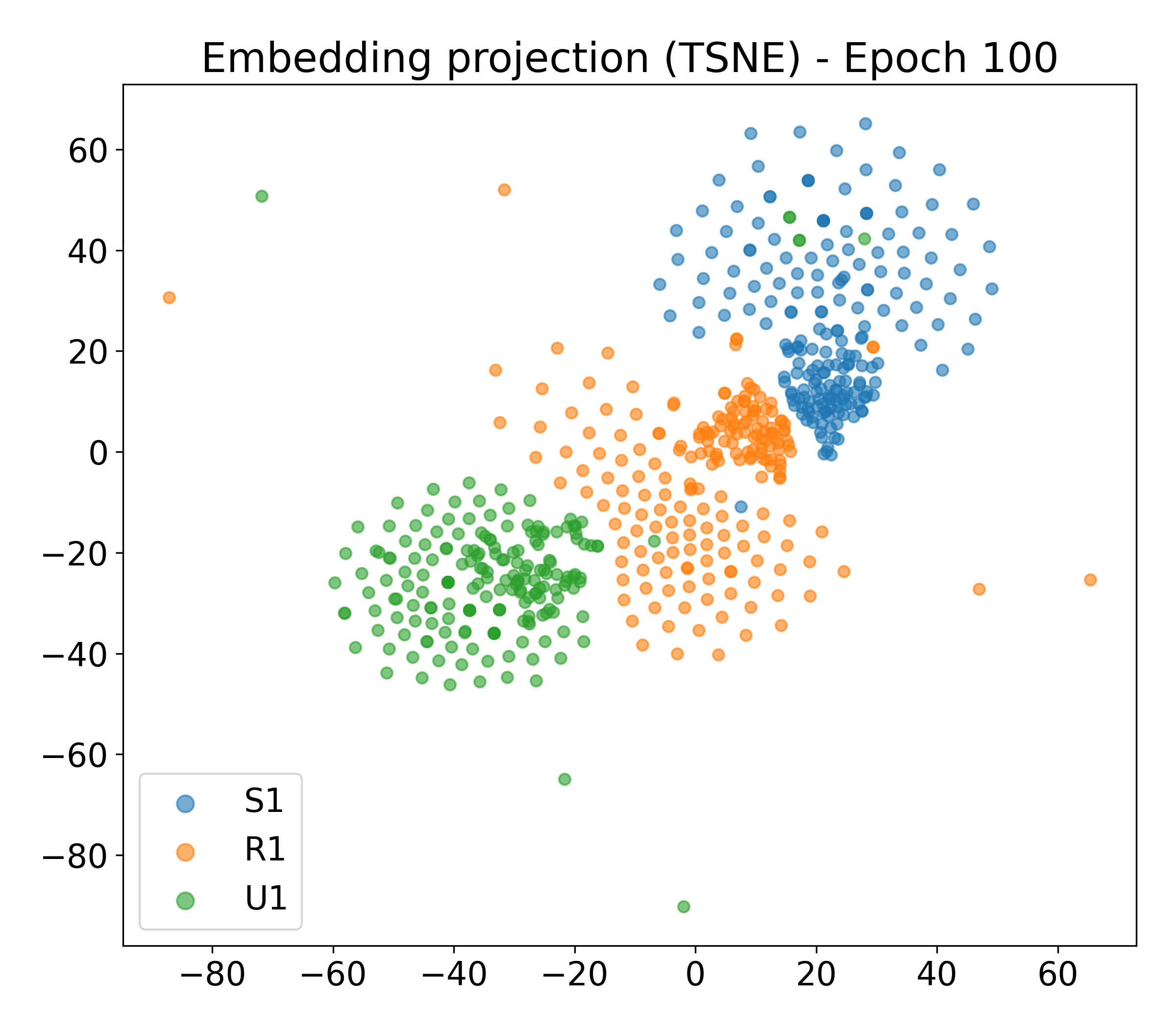} \\
    \end{tabular}
  }
  \caption{t-SNE visualization of the PID process embeddings at three training stages (epoch 0, 10, 100). Each subfigure shows the 2-D t-SNE projection; colors correspond to class/label if provided.}
  \label{fig:tsne_pid_epochs}
\end{figure*}

\subsection{Ablation Study}

\begin{figure}
    \centering
    \includegraphics[width=1\linewidth]{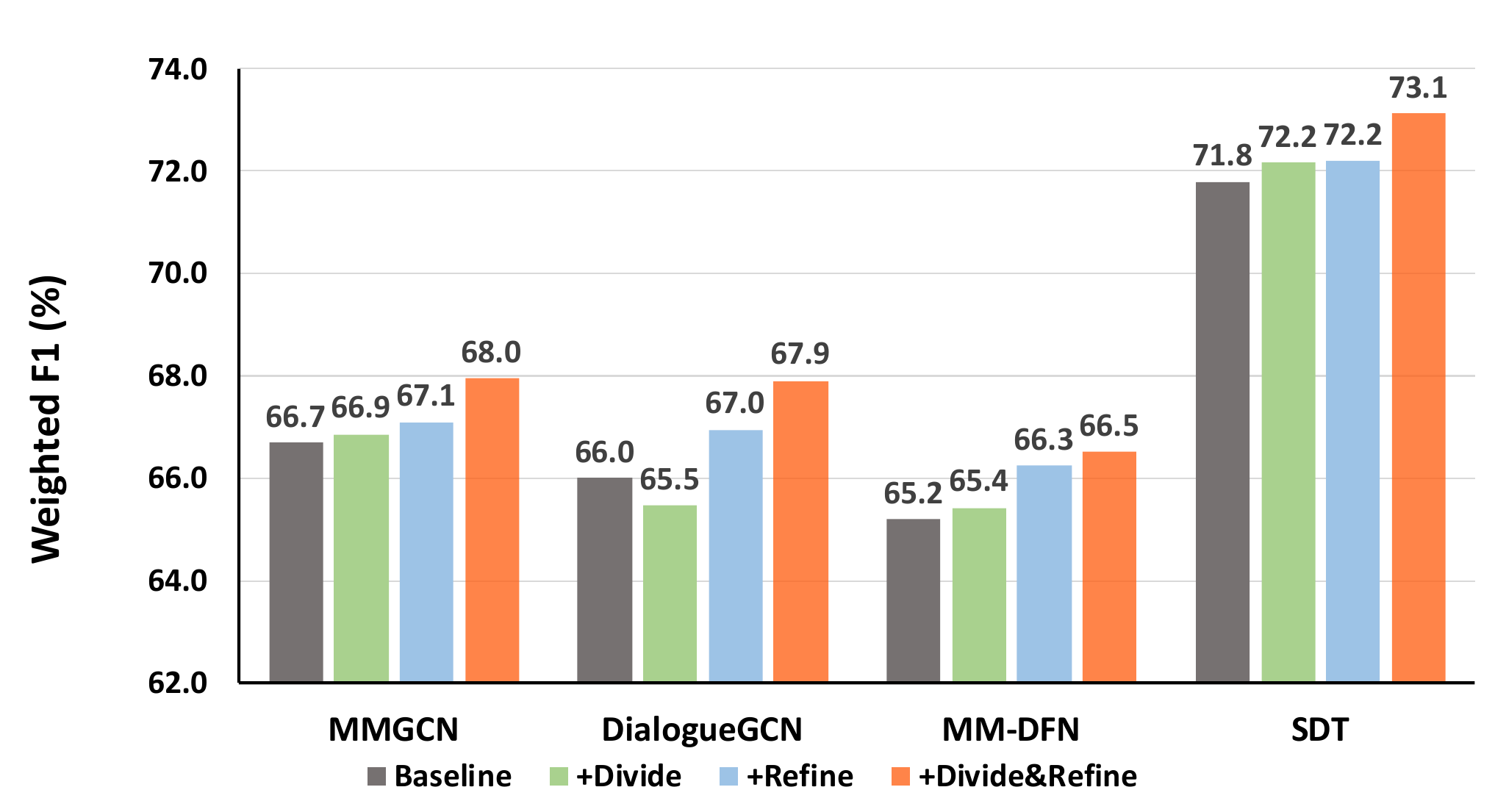}
    \caption{Ablation study of \mName{} conducted on IEMOCAP dataset.}
    \label{fig:module_ablation}
\end{figure}

To investigate the contributions of each phase in our framework, we perform an ablation study on the IEMOCAP dataset across all backbones under the \textit{atv} setting.  
We compare four settings namely: 
\begin{itemize}
    \item \textbf{Baseline models}
    \item \textbf{(+Refine)}: Decomposition is removed and augmentations are applied directly to raw input.
    \item \textbf{(+Divide)}: Decomposition is applied but the redundancy-refinement stage is omitted. 
    \item \textbf{(+Divide + Refine)}: Both phases are applied.
\end{itemize}
Results indicate that applying \textbf{Divide} alone does not always improve performance (e.g., while \textbf{Divide} helps in most cases, \textbf{DialogueGCN+Divide} drops in weighted-F1 from 66.01 to 65.46). In contrast, \textbf{Refine} consistently yields moderate gains, confirming its effectiveness in enhancing modality representation. Nevertheless, \textbf{Divide} remains a crucial foundation: combined with \textbf{Refine}, the model achieves the best results across backbones. This stems from a key \textbf{Refine} objective—maximizing mutual information between each modality-specific representation and the redundancy-perturbed representation. Without \textbf{Divide}, Gaussian-noise augmentation at the raw input perturbs redundancy and dampens each modality’s uniqueness, producing a suboptimal $Z$ even as mutual information between two noisy views increases. Figure~\ref{fig:module_ablation} presents the ablation study on IEMOCAP under the full-modality \textit{atv} setting.

\subsection{Decomposition Process}
Figure~\ref{fig:tsne_pid_epochs} shows the t-SNE visualization of the decomposition of a training textual modality sample. At the beginning (Epoch 0), different components are entangled. As training progresses (Epoch 10), clearer clusters emerge, and by Epoch 100, the unique, redundant, and synergistic parts are separated. The unique part (U1) exhibits the most distinct separation, while redundant (R1) and synergistic (S1) parts become increasingly organized.

\subsection{Qualitative Case Study}

\textbf{Case Study 1: Uniqueness–Redundancy Collapse.}  
This case highlights how the collapse of unique and redundant information can lead to misclassification. Utterance 31 (“Do you want to get married again? — What? Do you want a divorce?”) is labeled \emph{Sad}, but its uniqueness (of textual modality) and redundancy streams are highly overlapping, with $\text{KL}(m_1 \| m_1^{R}) = 0.016$, the lowest among the three utterances. From a PID perspective, this indicates that the utterance provides little unique information ($U$) beyond what is already redundantly encoded ($R$). In practice, such overlaps make it difficult for previous models, which easily ignore subtle unique information. 

\begin{table*}[!htbp]
\centering
\caption{Case study from IEMOCAP: three consecutive utterances (30--32). Utterance 31 shows a collapse between uniqueness and redundancy (low KL), which causes MMGCN to fail while our method succeeds.}
\resizebox{0.72\textwidth}{!}{%
\begin{tabular}{p{0.32\textwidth} p{0.32\textwidth} p{0.32\textwidth}}
\hline
\multicolumn{3}{c}{\textbf{Conversation snippet (utterances 30--32)}} \\ \hline

\begin{minipage}[t]{\linewidth}\centering
\includegraphics[width=0.6\linewidth]{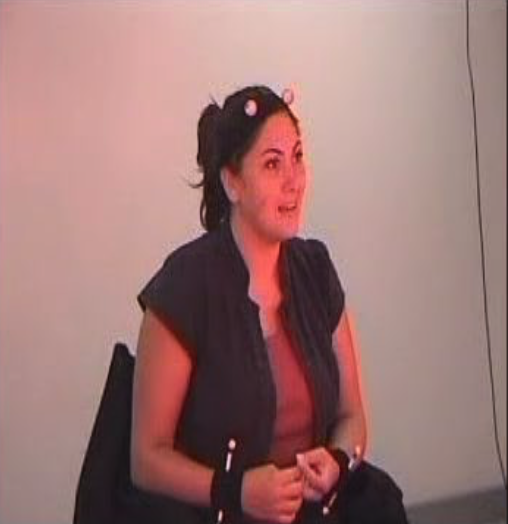}\\[4pt]
\textbf{Voice:} Slightly teasing, raised tone\\[4pt]
\textbf{Utterance:}\\
What, you want me to blow in your ear?\\[6pt]
\textbf{True:} Frustrated\\[4pt]
\textbf{MMGCN:} Frustrated\\[4pt]
\textbf{Ours:} Frustrated\\[6pt]
\textbf{KL$(m1||m1_{redundamt})=0.0525$}
\end{minipage}

&

\begin{minipage}[t]{\linewidth}\centering
\includegraphics[width=0.6\linewidth]{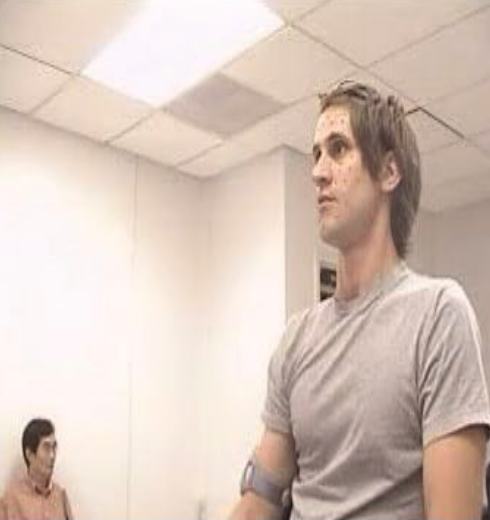}\\[4pt]
\textbf{Voice:} Neutral tone, flat delivery\\[4pt]
\textbf{Utterance:}\\
Do you want to get married again? --- What? Do you want a divorce?\\[6pt]
\textbf{True:} Sad\\[4pt]
\textbf{MMGCN:} Frustrated\\[4pt]
\textbf{Ours:} Sad\\[6pt]
{\textbf{KL$(m1||m1_{redundamt})=0.016$ (lowest)}}
\end{minipage}

&

\begin{minipage}[t]{\linewidth}\centering
\includegraphics[width=0.6\linewidth]{figures/frame4.png}\\[4pt]
\textbf{Voice:} Questioning, upward inflection\\[4pt]
\textbf{Utterance:}\\
A vacation?\\[6pt]
\textbf{True:} Neutral\\[4pt]
\textbf{MMGCN:} Neutral\\[4pt]
\textbf{Ours:} Neutral\\[6pt]
\textbf{KL$(m1||m1_{redundamt})=0.026$}
\end{minipage}

\\ \hline
\end{tabular}%
}
\end{table*}

\begin{table*}[!htbp]
\centering
\caption{Case study from IEMOCAP: four consecutive utterances with ground-truth \textbf{Sad}. 
KL divergence $\text{KL}(m_3 \| m_{3}^{R})$ becomes small for repeated ``No'' utterances (27--29), showing reduced uniqueness and making them difficult for baseline models.}
\resizebox{0.72\textwidth}{!}{%
\begin{tabular}{p{0.24\textwidth} p{0.24\textwidth} p{0.24\textwidth} p{0.24\textwidth}}
\hline
\multicolumn{4}{c}{\textbf{Conversation snippet (utterances 26--29)}} \\ \hline

\begin{minipage}[t]{\linewidth}\centering
\includegraphics[width=0.8\linewidth]{figures/frame3.png}\\[4pt]
\textbf{Voice:} Low, hesitant tone\\[2pt]
\textbf{Utterance:}\\
I would rather not remember some things. I'd rather not hope for something.\\[4pt]
\textbf{True:} Sad\\[4pt]
\textbf{MMGCN:} Neutral\\[2pt]
\textbf{Ours:} Sad\\[6pt]
\textbf{KL($m_3\|m_{3}^{R}$)} = 0.1602
\end{minipage}

&

\begin{minipage}[t]{\linewidth}\centering
\includegraphics[width=0.8\linewidth]{figures/frame4.png}\\[4pt]
\textbf{Voice:} Short, abrupt\\[2pt]
\textbf{Utterance:}\\
No.\\[4pt]
\textbf{True:} Sad\\[4pt]
\textbf{MMGCN:} Neutral\\[2pt]
\textbf{Ours:} Sad\\[6pt]
\textbf{KL($m_3\|m_{3}^{R}$)} = 0.0902
\end{minipage}

&

\begin{minipage}[t]{\linewidth}\centering
\includegraphics[width=0.8\linewidth]{figures/frame3.png}\\[4pt]
\textbf{Voice:} Same curt delivery\\[2pt]
\textbf{Utterance:}\\
No.\\[4pt]
\textbf{True:} Sad\\[4pt]
\textbf{MMGCN:} Neutral\\[2pt]
\textbf{Ours:} Sad\\[6pt]
\textbf{KL($m_3\|m_{3}^{R}$)} = 0.1016
\end{minipage}

&

\begin{minipage}[t]{\linewidth}\centering
\includegraphics[width=0.8\linewidth]{figures/frame4.png}\\[4pt]
\textbf{Voice:} Same curt delivery\\[2pt]
\textbf{Utterance:}\\
No.\\[4pt]
\textbf{True:} Sad\\[4pt]
\textbf{MMGCN:} Neutral\\[2pt]
\textbf{Ours:} Sad\\[6pt]
\textbf{KL($m_3\|m_{3}^{R}$)} = 0.1008
\end{minipage}

\\ \hline
\end{tabular}%
}
\end{table*}

\textbf{Case Study 2: Short utterance.}  
This case illustrates a PID-consistent challenge in multimodal emotion recognition: short utterances such as “NO” (utterances 27–29). From a PID perspective, these utterances contain minimal unique information ($U$), their meaning is largely redundant ($R$) with contextual cues, and they contribute little or no new synergistic information ($S$). The KL divergence between uniqueness (from visual modality) and redundancy vectors, $\text{KL}(m_3 \| m_3^{R})$, supports this observation: values for utterances 27–29 (0.09–0.10) are substantially lower than that of utterance 26 (0.16), indicating a collapse of $U$ and $R$ into near-indistinguishability. In such redundancy-dominated regimes, baseline MMGCN fails, defaulting to \emph{Neutral} because it cannot capture the subtle emotional signal when $U$ is weak.  

In contrast, our \textbf{\mName{}} correctly predicts \emph{Sad} across all four utterances. This robustness stems from its divide-and-refine design: decomposition disentangles $U$, $R$, and $S$, while refinement strengthens uniqueness via augmentation, ensuring trivial correlations do not overwhelm the emotional signal. Consequently, even when $U$ collapses, refined $R$ retains enough evidence to align with ground truth. This shows how \textbf{\mName{}} operationalizes PID in practice, mitigating failure modes where conventional models overlook subtle yet redundant emotional cues.

\section{Conclusion}

In this work, we presented a two-phase plug-and-play framework for multimodal emotion recognition in conversation, grounded in Partial Information Decomposition. Our approach disentangles modality representations into unique, redundant, and synergistic components (\textbf{Divide}), and selectively refines redundancy through representation-level augmentation (\textbf{Refine}). This design enhances the informativeness of multimodal features while remaining compatible with diverse MER backbones. Extensive experiments on IEMOCAP and MELD demonstrated consistent improvements over state-of-the-art baselines, and case studies further illustrated how the framework mitigates common failure modes where existing models conflate informational roles.
Future directions include extending Divide–and–Refine to other multimodal tasks such as sentiment analysis or intent recognition, and exploring adaptive refinement strategies that dynamically balance uniqueness, redundancy, and synergy based on conversational context.

\section*{Acknowledgement}
This research was partially funded by the research project QG.23.37 of Vietnam National University, Hanoi.

{
    \small
    \balance
    \bibliographystyle{ieeenat_fullname}
    \bibliography{main}

@article{ince2017measuring,
  title={Measuring multivariate redundant information with pointwise common change in surprisal},
  author={Ince, Robin AA},
  journal={Entropy},
  volume={19},
  number={7},
  pages={318},
  year={2017},
  publisher={MDPI}
}

@article{williams2010nonnegative,
  title={Nonnegative decomposition of multivariate information},
  author={Williams, Paul L and Beer, Randall D},
  journal={arXiv preprint arXiv:1004.2515},
  year={2010}
}

@article{bertschinger2014quantifying,
  title={Quantifying unique information},
  author={Bertschinger, Nils and Rauh, Johannes and Olbrich, Eckehard and Jost, J{\"u}rgen and Ay, Nihat},
  journal={Entropy},
  volume={16},
  number={4},
  pages={2161--2183},
  year={2014},
  publisher={Multidisciplinary Digital Publishing Institute}
}

@article{busso2008iemocap,
  title={IEMOCAP: Interactive emotional dyadic motion capture database},
  author={Busso, Carlos and Bulut, Murtaza and Lee, Chi-Chun and Kazemzadeh, Abe and Mower, Emily and Kim, Samuel and Chang, Jeannette N and Lee, Sungbok and Narayanan, Shrikanth S},
  journal={Language resources and evaluation},
  volume={42},
  number={4},
  pages={335--359},
  year={2008},
  publisher={Springer}
}

@inproceedings{wortwein2024smurf,
  title={Smurf: Statistical modality uniqueness and redundancy factorization},
  author={W{\"o}rtwein, Torsten and Allen, Nicholas B and Cohn, Jeffrey F and Morency, Louis-Philippe},
  booktitle={Proceedings of the 26th International Conference on Multimodal Interaction},
  pages={339--349},
  year={2024}
}

@article{liang2023factorized,
  title={Factorized contrastive learning: Going beyond multi-view redundancy},
  author={Liang, Paul Pu and Deng, Zihao and Ma, Martin Q and Zou, James Y and Morency, Louis-Philippe and Salakhutdinov, Ruslan},
  journal={Advances in Neural Information Processing Systems},
  volume={36},
  pages={32971--32998},
  year={2023}
}

@inproceedings{poria2018meld,
  title={MELD: A Multimodal Multi-Party Dataset for Emotion Recognition in Conversations},
  author={Poria, Soujanya and Hazarika, Devamanyu and Majumder, Navonil and Naik, Gautam and Cambria, Erik and Mihalcea, Rada},
  booktitle={Proceedings of the 57th Annual Meeting of the Association for Computational Linguistics},
  pages={527--536},
  year={2019}
}

@article{pan2023review,
  title={A review of multimodal emotion recognition from datasets, preprocessing, features, and fusion methods},
  author={Pan, Bei and Hirota, Kaoru and Jia, Zhiyang and Dai, Yaping},
  journal={Neurocomputing},
  volume={561},
  pages={126866},
  year={2023},
  publisher={Elsevier}
}

@article{liang2023quantifying,
  title={Quantifying \& modeling multimodal interactions: An information decomposition framework},
  author={Liang, Paul Pu and Cheng, Yun and Fan, Xiang and Ling, Chun Kai and Nie, Suzanne and Chen, Richard and Deng, Zihao and Allen, Nicholas and Auerbach, Randy and Mahmood, Faisal and others},
  journal={Advances in Neural Information Processing Systems},
  volume={36},
  pages={27351--27393},
  year={2023}
}

@inproceedings{mai2019divide,
  title={Divide, conquer and combine: Hierarchical feature fusion network with local and global perspectives for multimodal affective computing},
  author={Mai, Sijie and Hu, Haifeng and Xing, Songlong},
  booktitle={Proceedings of the 57th annual meeting of the association for computational linguistics},
  pages={481--492},
  year={2019}
}

@article{ahmed2023systematic,
  title={A systematic survey on multimodal emotion recognition using learning algorithms},
  author={Ahmed, Naveed and Al Aghbari, Zaher and Girija, Shini},
  journal={Intelligent Systems with Applications},
  volume={17},
  pages={200171},
  year={2023},
  publisher={Elsevier}
}

@article{liang2024foundations,
  title={Foundations \& trends in multimodal machine learning: Principles, challenges, and open questions},
  author={Liang, Paul Pu and Zadeh, Amir and Morency, Louis-Philippe},
  journal={ACM Computing Surveys},
  volume={56},
  number={10},
  pages={1--42},
  year={2024},
  publisher={ACM New York, NY}
}

@inproceedings{hu-etal-2021-mmgcn,
    title = "{MMGCN}: Multimodal Fusion via Deep Graph Convolution Network for Emotion Recognition in Conversation",
    author = "Hu, Jingwen  and
      Liu, Yuchen  and
      Zhao, Jinming  and
      Jin, Qin",
    editor = "Zong, Chengqing  and
      Xia, Fei  and
      Li, Wenjie  and
      Navigli, Roberto",
    booktitle = "Proceedings of the 59th Annual Meeting of the Association for Computational Linguistics and the 11th International Joint Conference on Natural Language Processing (Volume 1: Long Papers)",
    month = aug,
    year = "2021",
    address = "Online",
    publisher = "Association for Computational Linguistics",
    url = "https://aclanthology.org/2021.acl-long.440/",
    doi = "10.18653/v1/2021.acl-long.440",
    pages = "5666--5675",
}

@inproceedings{eyben2010opensmile,
  title={Opensmile: the munich versatile and fast open-source audio feature extractor},
  author={Eyben, Florian and W{\"o}llmer, Martin and Schuller, Bj{\"o}rn},
  booktitle={Proceedings of the 18th ACM international conference on Multimedia},
  pages={1459--1462},
  year={2010}
}

@inproceedings{baltrusaitis2018openface,
  title={Openface 2.0: Facial behavior analysis toolkit},
  author={Baltrusaitis, Tadas and Zadeh, Amir and Lim, Yao Chong and Morency, Louis-Philippe},
  booktitle={2018 13th IEEE international conference on automatic face \& gesture recognition (FG 2018)},
  pages={59--66},
  year={2018},
  organization={IEEE}
}

@article{baltruvsaitis2018multimodal,
  title={Multimodal machine learning: A survey and taxonomy},
  author={Baltru{\v{s}}aitis, Tadas and Ahuja, Chaitanya and Morency, Louis-Philippe},
  journal={IEEE transactions on pattern analysis and machine intelligence},
  volume={41},
  number={2},
  pages={423--443},
  year={2018},
  publisher={IEEE}
}

@inproceedings{zadeh2018memory,
  title={Memory fusion network for multi-view sequential learning},
  author={Zadeh, Amir and Liang, Paul Pu and Mazumder, Navonil and Poria, Soujanya and Cambria, Erik and Morency, Louis-Philippe},
  booktitle={Proceedings of the AAAI conference on artificial intelligence},
  volume={32},
  number={1},
  year={2018}
}

@inproceedings{shen2021dialogxl,
  title={Dialogxl: All-in-one xlnet for multi-party conversation emotion recognition},
  author={Shen, Weizhou and Chen, Junqing and Quan, Xiaojun and Xie, Zhixian},
  booktitle={Proceedings of the AAAI Conference on Artificial Intelligence},
  volume={35},
  number={15},
  pages={13789--13797},
  year={2021}
}

@article{lian2022smin,
  title={Smin: Semi-supervised multi-modal interaction network for conversational emotion recognition},
  author={Lian, Zheng and Liu, Bin and Tao, Jianhua},
  journal={IEEE Transactions on Affective Computing},
  volume={14},
  number={3},
  pages={2415--2429},
  year={2022},
  publisher={IEEE}
}

@inproceedings{ghosal2019dialoguegcn,
    title = "{D}ialogue{GCN}: A Graph Convolutional Neural Network for Emotion Recognition in Conversation",
    author = "Ghosal, Deepanway  and
      Majumder, Navonil  and
      Poria, Soujanya  and
      Chhaya, Niyati  and
      Gelbukh, Alexander",
    booktitle = "Proceedings of the 2019 Conference on Empirical Methods in Natural Language Processing",
    month = nov,
    year = "2019",
    address = "Hong Kong, China",
    publisher = "Association for Computational Linguistics",
    doi = "10.18653/v1/D19-1015",
    pages = "154--164",
}

@inproceedings{ghosal2020cosmic,
    title = "{COSMIC}: {CO}mmon{S}ense knowledge for e{M}otion Identification in Conversations",
    author = "Ghosal, Deepanway  and
      Majumder, Navonil  and
      Gelbukh, Alexander  and
      Mihalcea, Rada  and
      Poria, Soujanya",
    editor = "Cohn, Trevor  and
      He, Yulan  and
      Liu, Yang",
    booktitle = "Findings of the Association for Computational Linguistics: EMNLP 2020",
    month = nov,
    year = "2020",
    address = "Online",
    publisher = "Association for Computational Linguistics",
    url = "https://aclanthology.org/2020.findings-emnlp.224/",
    doi = "10.18653/v1/2020.findings-emnlp.224",
    pages = "2470--2481",
}

@inproceedings{nguyen-etal-2024-curriculum,
    title = "Curriculum Learning Meets Directed Acyclic Graph for Multimodal Emotion Recognition",
    author = "Nguyen, Cam-Van Thi  and
      Nguyen, Cao-Bach  and
      Le, Duc-Trong  and
      Ha, Quang-Thuy",
    booktitle = "Proceedings of the 2024 Joint International Conference on Computational Linguistics, Language Resources and Evaluation (LREC-COLING 2024)",
    month = may,
    year = "2024",
    address = "Torino, Italia",
    publisher = "ELRA and ICCL",
    pages = "4259--4265",
}

@inproceedings{nguyen2023conversation,
    title = "Conversation Understanding using Relational Temporal Graph Neural Networks with Auxiliary Cross-Modality Interaction",
    author = "Nguyen, Cam-Van Thi  and
      Mai, Anh-Tuan  and
      Le, The-Son  and
      Kieu, Hai-Dang  and
      Le, Duc-Trong",
    booktitle = "Proceedings of the 2023 Conference on Empirical Methods in Natural Language Processing",
    month = dec,
    year = "2023",
    address = "Singapore",
    publisher = "Association for Computational Linguistics",
    doi = "10.18653/v1/2023.emnlp-main.937",
    pages = "15154--15167",
}

@inproceedings{hu2021dialoguecrn,
    title = "{D}ialogue{CRN}: Contextual Reasoning Networks for Emotion Recognition in Conversations",
    author = "Hu, Dou  and
      Wei, Lingwei  and
      Huai, Xiaoyong",
    booktitle = "Proceedings of the 59th Annual Meeting of the Association for Computational Linguistics and the 11th International Joint Conference on Natural Language Processing (Volume 1: Long Papers)",
    month = aug,
    year = "2021",
    address = "Online",
    publisher = "Association for Computational Linguistics",
    doi = "10.18653/v1/2021.acl-long.547",
    pages = "7042--7052",
    
}

@article{tao2021short,
  title={Short-term forecasting of photovoltaic power generation based on feature selection and bias compensation--LSTM network},
  author={Tao, Cai and Lu, Junjie and Lang, Jianxun and Peng, Xiaosheng and Cheng, Kai and Duan, Shanxu},
  journal={Energies},
  volume={14},
  number={11},
  pages={3086},
  year={2021},
  publisher={MDPI}
}

@inproceedings{hazarika2018conversational,
  title={Conversational memory network for emotion recognition in dyadic dialogue videos},
  author={Hazarika, Devamanyu and Poria, Soujanya and Zadeh, Amir and Cambria, Erik and Morency, Louis-Philippe and Zimmermann, Roger},
  booktitle={Proceedings of the conference. Association for Computational Linguistics. North American Chapter. Meeting},
  volume={2018},
  pages={2122},
  year={2018}
}

@article{ma2023transformer,
  title={A transformer-based model with self-distillation for multimodal emotion recognition in conversations},
  author={Ma, Hui and Wang, Jian and Lin, Hongfei and Zhang, Bo and Zhang, Yijia and Xu, Bo},
  journal={IEEE Transactions on Multimedia},
  volume={26},
  pages={776--788},
  year={2023},
  publisher={IEEE}
}

@article{li2025tracing,
  title={Tracing intricate cues in dialogue: Joint graph structure and sentiment dynamics for multimodal emotion recognition},
  author={Li, Jiang and Wang, Xiaoping and Zeng, Zhigang},
  journal={IEEE Transactions on Pattern Analysis and Machine Intelligence},
  year={2025},
  publisher={IEEE}
}

@inproceedings{joshi2022cogmen,
    title = "{COGMEN}: {CO}ntextualized {GNN} based Multimodal Emotion recognitio{N}",
    author = "Joshi, Abhinav  and
      Bhat, Ashwani  and
      Jain, Ayush  and
      Singh, Atin  and
      Modi, Ashutosh",
    
    booktitle = "Proceedings of the 2022 Conference of the North American Chapter of the Association for Computational Linguistics: Human Language Technologies",
    month = jul,
    year = "2022",
    address = "Seattle, United States",
    publisher = "Association for Computational Linguistics",
    doi = "10.18653/v1/2022.naacl-main.306",
    pages = "4148--4164",
}

@inproceedings{dufumier2024align,
  title={What to align in multimodal contrastive learning?},
  author={Dufumier, Benoit and Navarro, Javiera Castillo and Tuia, Devis and Thiran, Jean-Philippe},
  booktitle={The Thirteenth International Conference on Learning Representations}
}

@inproceedings{song2025leveraging,
  title={Leveraging CLIP Encoder for Multimodal Emotion Recognition},
  author={Song, Yehun and Cho, Sunyoung},
  booktitle={2025 IEEE/CVF Winter Conference on Applications of Computer Vision (WACV)},
  pages={6115--6124},
  year={2025},
  organization={IEEE}
}

@inproceedings{reimers-gurevych-2019-sentence,
    title = "Sentence-{BERT}: Sentence Embeddings using {S}iamese {BERT}-Networks",
    author = "Reimers, Nils  and
      Gurevych, Iryna",
    booktitle = "Proceedings of the 2019 Conference on Empirical Methods in Natural Language Processing and the 9th International Joint Conference on Natural Language Processing (EMNLP-IJCNLP)",
    month = nov,
    year = "2019",
    address = "Hong Kong, China",
    publisher = "Association for Computational Linguistics",
    doi = "10.18653/v1/D19-1410",
    pages = "3982--3992"
}

@article{schneider2019wav2vec,
  title={wav2vec: Unsupervised pre-training for speech recognition},
  author={Schneider, Steffen and Baevski, Alexei and Collobert, Ronan and Auli, Michael},
  journal={arXiv preprint arXiv:1904.05862},
  year={2019}
}

@article{zhao2021learning,
  title={Learning deep global multi-scale and local attention features for facial expression recognition in the wild},
  author={Zhao, Zengqun and Liu, Qingshan and Wang, Shanmin},
  journal={IEEE Transactions on Image Processing},
  volume={30},
  pages={6544--6556},
  year={2021},
  publisher={IEEE}
}

@article{zhang2016joint,
  title={Joint face detection and alignment using multitask cascaded convolutional networks},
  author={Zhang, Kaipeng and Zhang, Zhanpeng and Li, Zhifeng and Qiao, Yu},
  journal={IEEE signal processing letters},
  volume={23},
  number={10},
  pages={1499--1503},
  year={2016},
  publisher={IEEE}
}

@article{he2006deberta,
  title={Deberta: Decoding-enhanced bert with disentangled attention. arXiv 2020},
  author={He, Pengcheng and Liu, Xiaodong and Gao, Jianfeng and Chen, Weizhu},
  journal={arXiv preprint arXiv:2006.03654},
  year={2006}
}

@inproceedings{hu2022mm,
  title={MM-DFN: Multimodal dynamic fusion network for emotion recognition in conversations},
  author={Hu, Dou and Hou, Xiaolong and Wei, Lingwei and Jiang, Lianxin and Mo, Yang},
  booktitle={ICASSP 2022-2022 IEEE International Conference on Acoustics, Speech and Signal Processing (ICASSP)},
  pages={7037--7041},
  year={2022},
  organization={IEEE}
}
}

\end{document}